\documentclass[journal]{IEEEtran}
\usepackage{pgfplots}
\pgfplotsset{width=8cm,compat=1.16}

\usepackage{cite}
\usepackage{amsmath,amssymb,amsfonts}
\usepackage{algorithmic}
\usepackage{graphicx}
\usepackage{textcomp}
\usepackage{float}
\usepackage{caption}
\usepackage{subcaption}

\begin{document}
\title{Noise Conscious Training of Non Local Neural Network powered  by Self Attentive Spectral Normalized Markovian Patch GAN for Low Dose CT Denoising}
\author{Sutanu Bera, Prabir Kumar Biswas, \IEEEmembership{Senior Member, IEEE}

\thanks{Sutanu Bera, is with Department of Electronics and Electrical Communication Engineering, Indian Insitute of Techonology Kharagpur. (e-mail: sutanu.bera@iitkgp.ac.in).}
\thanks{Prabir Kumar Biswas, is with Department of Electronics and Electrical Communication Engineering, Indian Insitute of Techonology Kharagpur. (e-mail: pkb@ece.iitkgp.ac.in)}}

\maketitle

\begin{abstract}
 The explosive rise of the use of Computer tomography (CT) imaging in medical practice has heightened public concern over the patient's associated radiation dose. However, reducing the radiation dose leads to increased noise and artifacts, which adversely degrades the scan's interpretability. Consequently, an advanced image reconstruction algorithm to improve the diagnostic performance of low dose ct arose as the primary concern among the researchers, which is challenging due to the ill-posedness of the problem. In recent times, the deep learning-based technique has emerged as a dominant method for low dose CT(LDCT) denoising. However, some common bottleneck still exists, which hinders deep learning-based techniques from furnishing the best performance. In this study, we attempted to mitigate these problems with three novel accretions. First, we propose a novel convolutional module as the first attempt to utilize neighborhood similarity of CT images for denoising tasks. Our proposed module assisted in boosting the denoising by a significant margin. Next, we moved towards the problem of non-stationarity of CT noise and introduced a new noise aware mean square error loss for LDCT denoising. Moreover, the loss mentioned above also assisted to alleviate the laborious effort required while training CT denoising network using image patches. Lastly, we propose a novel discriminator function for CT denoising tasks. The conventional vanilla discriminator tends to overlook the fine structural details and focus on the global agreement. Our proposed discriminator leverage self-attention and pixel-wise GANs for restoring the diagnostic quality of LDCT images. Our method validated on a publicly available dataset of the 2016 NIH-AAPM-Mayo Clinic Low Dose CT Grand Challenge performed remarkably better than the existing state of the art method.\\
\end{abstract}

\section{Introduction}
Computerized Tomography (CT) has enabled direct imaging of the 3-dimensional structure of different organs and tissues inside the human body in a non-invasive manner. It is constructed by combining
the X-Ray scans taken at different angles and orientations. CT scan has several utilities but is especially useful in detecting lesions,
tumors, and metastasis. It can reveal their presence and the spatial location, size, and extent of the tumor. In modern days, CT imaging has become a frequent tool for cancer diagnosis, angiography, and
detecting internal injuries. However, despite the evidence of the utility of CT scan for diagnosis and patient management, the potential risk of radiation-induced malignancy nevertheless exists [1]. Studies found that CT alone contributes almost one half of the total radiation. exposure from medical use [2]. Recent studies reveal that as much as $1.5 - 2 \%$ of cancers may eventually be caused by the radiation dose conceived by the patient while doing CT [3]. To overwhelm the adverse effects, the principles of ALARA (As Low As Reasonably Achievable) is now profoundly practiced in CT imaging [4]. In this regard, Low Dose CT (LDCT) is a promising key. [5]. In low dose CT, radiation exposure is decreased by lowering the tube current, or voltage. However, reducing the tube voltage or current introduces several artifacts, thus lowering the diagnostic quality of the LDCT image [6]. To boost the diagnostic quality, better methods for the reconstruction of LDCT images became primary research concerns. These methods can be broadly classified into three categories: (a) sinogram filtration based technique, [7], [8] (b) iterative reconstruction [9], [10], and (c) image post-processing based technique [11], [12].
Over the past decade, researchers were dedicated to developing new iterative algorithms (IR) for LDCT image reconstruction. The main principle of iterative reconstruction based technique is to optimize an objective function that incorporates an accurate system model\cite{ref12}, a statistical noise model\cite{ref13}, and prior information about the image. Different IR techniques adopt different prior assumptions, e.g., total variation (TV) and its variants\cite{ref14}, dictionary learning\cite{ref15}. These iterative reconstruction algorithms considerably suppressed the image noise, but still lose some details and suffer from remaining artifacts. Also, they demand a high computational facility, which is a bottleneck in practical utilization. 
On the other hand, sinogram filtration directly works on the projection data before reconstructing the image. These methods are computationally economical compared to iterative reconstruction.  However, sinogram data of commercial scanners are not readily available to users, and also these methods suffer from edge blurring and the resolution loss. Additionally, different artifacts emerged in the reconstructed images if sinogram data is not carefully processed. Conversely, image post-processing directly operates on the image domain. Many efforts were made in the image domain to reduce LDCT noise and suppress artifacts. Specifically, the non-local means-based method has received notable appreciation for their powerful denoising results\cite{ref_article5}. 
\par With the recent explosive evolution of deep neural networks, like any other image processing task, the LDCT denoising task is also now dominated by deep neural networks. For example, H Chen \textit{et al.}\cite{ref_article6} proposed a deep convolutional neural network named RED-CNN, for LDCT denoising. Their method outperformed all existing traditional image processing methods. Other deep neural network-based methods that worth mentioning are \cite{ref_article7}, \cite{ref_article8}, \cite{ref_article9}. However, the research of deep learning-based LDCT denoising is confined to designing network architecture based on the vanilla convolution operation. There is a scarcity of research in designing a new convolutional module, which can be incorporated in any architecture for intensified denoising performance. On the other hand, there has been a surge in interest in designing non local neural networks, mainly for object classification and detection in both medical images and natural images. Still, there is a dearth of literature for efficient non local networks concerned with image restoration despite the fact, non-local means based algorithms like BM3D\cite{ref_article10}, NLM\cite{ref_article11}, etc. have performed remarkably for medical image denoising. In this study, we are proposing a novel non-local module for denoising of LDCT images. Our proposed module uses self-similarity present in the neighborhood for enhanced denoising similar to classical nonlocal means methods. To our best knowledge, this is the first realization of a module utilizing neighborhood similarity for medical image denoising. Our proposed module has helped in achieving 0.4db more PSNR than the baseline networks. Also, our proposed network employing our proposed module produced denoising results had defeated all the state of art methods.
\par Another important aspect of CNN based method is training the network.  Most of the existing studies directly used per-pixel loss like MSE, MAE, etc. for training their network. However, CT noise is a non-uniform signal-dependent noise. So, some region image has more impact of noise, while some region with low signal intensity is less affected by noise. Considering this fact, giving equal weightage to every pixel in the image for calculating loss is nonsensical. In this study, we propose to use a novel noise aware MSE loss as a per-pixel loss. Our proposed loss function reasonably improved denoising results. This loss not only gives special attention to the regions with high signal intensity but also solves one salient issue of training CNN with CT image patches. As CT image patches contain air in a significant area, training CNN with those patches is cumbersome. Previous researchers [13] have tried manually to exclude image patches that were mostly air. Our proposed loss is also useful in this regard as it reduces the weightage of the area with air. With our proposed loss function, we have achieved 0.2db more PSNR than the same network trained with conventional MSE loss. 
But even so, reducing per-pixel loss often results in over-smoothed edges and loss of fine details, as reported in recent work\cite{ref_article14}. As a solution, Qingsong Yang \textit{et al.}\cite{ref_article15}  proposed to use adversarial loss coupled with a perceptual loss for LDCT denoising. The network trained with adversarial loss produces visually sharp images. Following, \cite{ref_article15} other researchers like H Shan \textit{et al.}\cite{ref_article16}, Chenyu You \textit{et al.}\cite{ref_article17} also used adversarial loss for training their network. The main principle of adversarial loss is to penalize an unrealistic generated image. It uses a discriminator or some time called a critic network to judge whether the reconstructed image is realistic or not. The generator network tries to fool the discriminator or critic by generating a better-looking, sharp, realistic image. It is noteworthy to point out that discriminator used in the existing studies in LDCT denoising classified/judged the entire image as a single object. However, CT images contain many small lesions, blood vessels, which may become less important for discriminator while classifying the whole image. As an alternative approach, many researchers tried to combine local discriminator with global discriminator to judge local regions separately.
In recent work, Jiahui Yu \textit{et al.}\cite{ref_article18} proposed a novel discriminator named Spectral-Normalized Markovian Patch (SNMP) Discriminator. The above-mentioned discriminator classifies each spatial location of input image separately, by formulating several standalone local GANs. Although the discriminator above performed well for a problem where distortion is more severe only in a local region (e.g., image inpainting), but lacks global consistency. Global consistency is a crucial requirement for LDCT denoising. In this study, we propose a self-attentive Spectral Normalized Markovian Patch Discriminator to ease the problem of the global consistency of the existing SNMP discriminator. We evaluated proposed discriminator performance by comparing contrast to noise ratio (CNR) of the different regions with lesion and anomalies with other methods and additionally by comparing Fréchet Inception Distance(FID) between original high dose CT images and restored images. Our discriminator defeated all previous methods in all evaluation metrics. 
\par In summary, we present the following technical contributions specifically for LDCT denoising in this paper 
\begin{itemize}
    \item A novel non local module exploiting neighbourhood similarity in CT images for enhanced denoising. 
    \item We introduced a novel MSE loss for LDCT denoising, which gives special attention to high signal intensity region of CT image, also mitigate the laborious work of training deep learning network with image patches.
    \item We proposed a discriminator network for training CNN generator network which gives special focus on small details present on the CT images.
\end{itemize}
The rest of the paper is organized as follows: Section II explains the proposed method. Section III specifies the experimental settings and specifies the material used in conducting this study, Section IV provides the evaluation strategy used to assess proposed method, The experimental result is presented and discussed in Section V,. Finally, the Section
V provides a retrospective of what was accomplished through this proposed novel approach.
\section{Method}
\subsection{Non Local Module}
\subsubsection{General Framework}
The generic non-local mean operation for deep neural network is defines as\cite{ref_article12}.
\begin{center}
    $y_i=\frac{1}{C(x)}\sum_{\forall j} f(x_i,x_j)g(x_j)$
\end{center}
Here $x_i$ is the input feature, and $y_i$ is the output response at the index $i$, and $j$ is the index that enumerates all possible positions. The function g computes a representation of the input signal at the position j. $C(x)$ is the normalization factor, and $f$ is a pairwise function, which calculates the similarity between i and all j. 
\subsubsection{Our Proposed Module}
The aforementioned framework used features from whole image to calculate output response of any feature at location $i$. However, we argue that combining features from all over the images hinders the network from determining the correct correlation of different features with the current feature; CT images contain different type of tissue at distant spatial location, and also significant segment of the CT image is air. So, instead of using the whole image we propose to use features from a small neighborhood surrounding the current feature to compute the response. Based on the above property, our Non local module is given by:
\begin{center}
    $y_i=\frac{1}{C(x)}\sum_{\forall j\in N_i} f(x_i,x_j)g(x_j)$
\end{center}
For computing the pair wise distance between features we used embedded Gaussian as given in\cite{ref_article12}. Following their formulation we also used:
\begin{center}
    $f(x_i,x_j)= e^{\theta(x_i)^T}e^{\phi(x_j)}$ and $y_i=\frac{1}{C(x)}\sum_{\forall j\in N_i} e^{\theta(x_i)^T} e^{\phi(x_j)}g(x_j)$
\end{center}

\begin{figure*}
    \centering
    \includegraphics[scale =0.4]{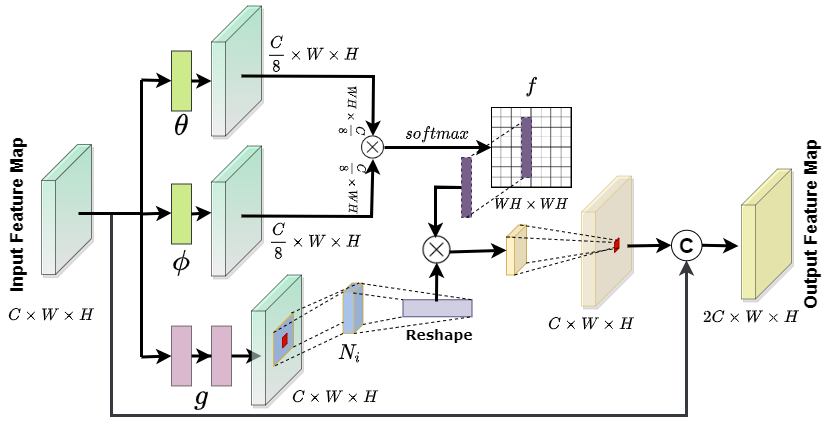}
    \caption{An illustration of our non-local module working on a single location. The stated location is highlighted in red. This module can be embedded in any convolutional neural network architecture. }
    \label{fig:nonL}
\end{figure*}
Here, $\theta$, $\phi$ are the two embedding of input feature map. We set $C(x)=\sum_{\forall j\in N_i}f(x_i,x_j)$.
So, for single pixel location i the response $y_i$ becomes.
\begin{center}
     $y_i=\frac{1}{\sum_{\forall j\in N_i}f(x_i,x_j)}\sum_{\forall j\in N_i} e^{\theta(x_i)^T} e^{\phi(x_j)}g(x_j)$
\end{center}
Next important part is how to fuse this feature with previous features. Previous researcher\cite{ref16},\cite{ref17} used residual connection for adding these features with the previous features, i.e. $z_i = \gamma * y_i + x_i$. They set the initial value of $\gamma$ as $0$, so the network first learn the local behaviour and slowly starts to learn non local behaviour. However we found in our case, the $\gamma$ actually converged to a negative value, which implies the network is learning negative non local behaviour. So, we forced the network to jointly learn the the non local and local behaviour by concatenating these non local features with previous features.

It is self explanatory that taking normalization over embedded Gaussian is equivalent to performing softmax operation on the dot product of the two embedding over the index $j$\cite{ref_article12}. We parameterized  $\phi$ and $\theta$ with single layer convolutional block of $(1 \times 1)$ kernel, and $g$ with 2 layer convolutional network with $(3 \times 3)$ kernel. We found that increasing the kernel size for $g$ helps in increasing performance. So, we used multi layer convolutional block for realising $g$. Our proposed nonlocal module is shown in Figure 1. 
\paragraph{Relationship with other module}
Recently, Meng Li \textit{et al.}\cite{ref_article31} proposed a self attention module for low dose CT denoising, however our module is not self attention as we have used only a local neighborhood. In self attention response is accumulated from all over the image, which is equivalent to expanding neighborhood size to whole image. In Section 4 we have discussed how the denoising performance depends on the size of neighbourhood. 
\paragraph{Difference with Vanilla Convolution Layer}
Our proposed module only used a small neighbourhood for computing the non local response. This analogues to vanilla convolution layer which also capture information with in the receptive field of every pixel. The operation of vanilla convolution layer is $y_i = \sum_{\forall j\in N_i} W_jx_j$. So $W_j$ is similar to $f(x_i,x_j)$ and $g()$ can be considered as identity mapping. However in vanilla convolution layer every pixel shares the same $W_j$, so it considers all pixels have similar correlation with their neighbours, however our proposed module calculates the correlation between every pixel with their neighbour. 
\subsection{Generator Network}
In this work a modified version of RED-CNN architecture is used as  generator. The nonlocal module is computationally expensive; the number of computations increases with the size of the input image. So, for CT images with $512 \times 512$ size, it can not be used with full image resolution, hence down sampling is essential. We used the max pool for down sampling the image. And, for up sampling, we used nearest-neighbor interpolation, followed by a standard $5 \times 5$ convolution layer. Here also features from the initial layer of the network are forwarded to the succeeding layer of the network in a symmetrical manner, similar to RED-CNN. Our non local module is implanted in the middle layer of the network, followed by the standard convolution layer. We set number of feature maps to 64 and size of kernel to $5 \times 5$ for every convolution layer. In Figure 2a our proposed generator is shown.
\begin{figure*}
    \centering
    \begin{subfigure}{\textwidth}
    \includegraphics[scale=0.5]{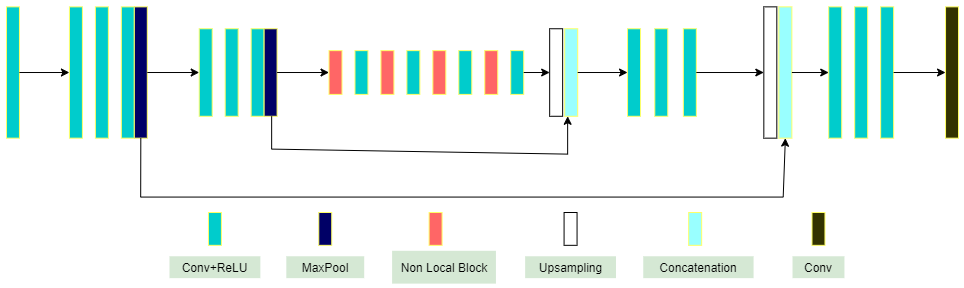}
    \caption{Generator Network}
    \end{subfigure}
    \begin{subfigure}{\textwidth}
    \includegraphics[scale=0.5]{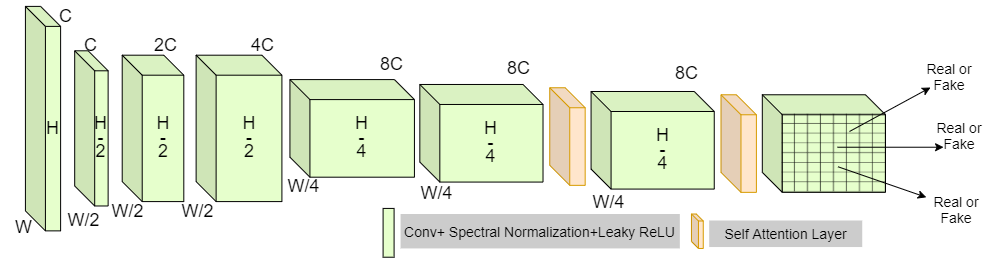}
    \caption{Discriminator Network}
    \end{subfigure}
    \caption{Illustration of our proposed Generator and Discriminator network. Viewers are encouraged to zoom for better view.}
    \label{fig:my_label}
\end{figure*}
\subsection{Discriminator Network}
We used self-attentive spectral normalized markovian patch(SA-SNMP) discriminator for training our generator network. Here a fully convolutional network is used as a discriminator; the output of the discriminator is a 3D feature map of size $\frac{h}{4} \times \frac{w}{4} \times c$, where the size of the input image is $h \times w$. This discriminator is inspired by [12], which was primarily introduced for image inpainting tasks. Compared with vanilla discriminator, SNMP discriminator has a significant advantage of focusing on each of the different locations and semantics separately. Radically, it is alike to apply several local discriminators in every location of the image. So, every local discriminator will judge a particular region of the image and guide it to generate a more realistic image. However, we found one major disadvantage of this discriminator, the lack of global consistency.
Most importantly, for the LDCT denoising task, global consistency is desired and alongside focusing on every small detail. The two easy solutions are as follows: \textbf{i.} By adding a global discriminator for training, \textbf{ii.} By adding more layers to the discriminator network. So that the receptive filed of every pixel in the final layer can cover the whole image. However, both the solution will lead to more trainable parameters hence unstable training. We combat this problem by using self-attention mechanism in the final layers of our discriminator network. As discussed in the previous section, self-attention is similar to our non-local module, except the output response is computed by combining the correlation from the whole image. So it is defined as 
\begin{center}
     $y_i=\frac{1}{\sum_{\forall j}f(x_i,x_j)}\sum_{\forall j} e^{\theta(x_i)^T} e^{\phi(x_j)}g(x_j)$
\end{center}
In this case we used the residual connection to add non local features to  original features.
Self attention can significantly increase the receptive field. In Figure 2b our proposed discriminator is shown. GANs are directly applied to each feature element in the final feature map, formulating $\frac{w}{4} \times \frac{h}{4} \times c$ number of GANs. However, unlike [12],  hinge loss is not used for training. Instead of conventional Wasserstein distance is used as an objective function for training as it more stable for LDCT denoising and also converged faster. So accordingly, the objective function used for training our generator network is
\begin{center}
$\mathcal{L}_{GA}=-\mathbb{E}_{z \backsim \mathbb{P}_z(z)}[D(G(z))]$
\end{center}
and objective function for discriminator is
\begin{center}
 $\mathcal{L}_D=-\mathbb{E}_{x \backsim \mathbb{P}_{data}(x)}[D(x)]+\mathbb{E}_{z \backsim \mathbb{P}_z(z)}[D(G(z))]$.
\end{center}

Where, $G$, and $D$ are the generator and discriminator respectively.
\subsection{Noise aware MSE Loss:}
 \begin{figure}[h]
     \centering
     \includegraphics[width=.48\textwidth]{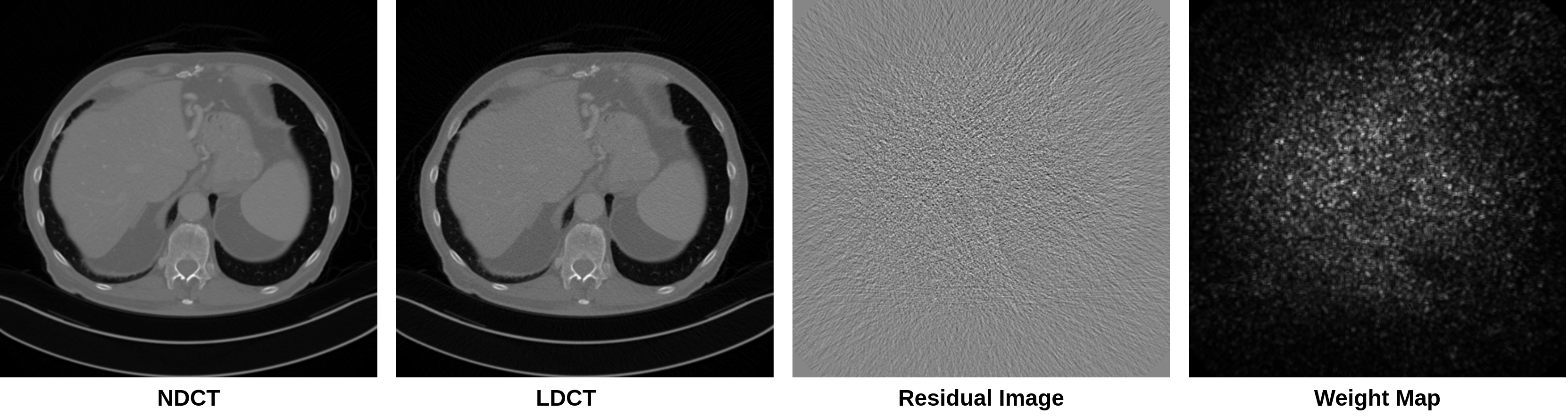}
     \caption{Predicted weight map by our method.Here it can be seen, the weightage of region filled with air is relatively low compared to the region with high signal intensity.}
     \label{weightmap}
 \end{figure}
Typically deep neural network uses Mean Square Error (MSE) loss to minimize the pixel wise distance between original noise less image patch and reconstructed image patch. The MSE loss is given by,
\begin{center}
    $L_{MSE}=\frac{1}{N}\sum_x\sum_y[I(x,y)-R(x,y)]^2$
\end{center}
Now, CT images patch contain air in a significant area; the network can learn to reconstruct those regions as there is less ambiguity. So, after a few epoch of training, the $L_{MSE}$ will be very small for all those image patches, and gradient update will be minimal. So, after a few epochs, practical training will stop implicitly. Another important note, CT noise is non-uniform signal-dependent. Noise affects more in the region where the signal intensity is high, and less in the area where the signal intensity is low. The network can learn to reconstruct the region with a low level of noise injection quickly, then again, the minimal gradient update will happen as a result of small $L_{MSE}$. We mitigate these two problems of training the deep neural network with CT image patches by applying noise-conscious MSE loss as an objective function. Our noise aware MSE loss is given by 
\begin{center}
   $L_{Gp}=\frac{1}{N}\sum_x\sum_y p(x,y)\odot [I(x,y)-R(x,y)]^2$
\end{center}
Here, $R(x,y)$ is reconstructed patch, $I(x,y)$ is the  normal dose CT (NDCT) patch, and $p(x,y)$ is the noise aware weight of the pixel $(x,y)$. Where, $p(x,y)$ is given by
\begin{center}
\begin{small}
 $p(x,y)=\sigma\left(\sqrt{\frac{1}{M}\sum_{(x,y) \in N_g} \left(G(x,y)-\frac{1}{M}\sum_{(x,y) \in N_g} G(x,y)\right)^2}\right)$.   
 \end{small}
\end{center}
Here, $\sigma ()$ is softmax operation given by $\sigma(x_i) = \frac{exp(x_i)}{\sum_{j}^{ }exp(x_j))}$. And $G(x,y)$ is given by 
\begin{center}
    $G(x,y)=W_g^{(x,y)}\odot \left(I(x,y)-R(x,y)\right)$
\end{center}
Here, $W_g$ is a rectangular sliding 2d Gaussian window of dimension $(h_g \times h_g)$ and standard deviation $\sigma_g$ placed at pixel location of $(x,y)$, and $N_g$ is all pixel inside the window $W_g$, and $\odot$ indicates element wise product. In Figure \ref{weightmap} weight map of our noise conscious loss is shown. Residual image is also shown for reference.The area filled with air has lowest weight, and region inside the liver has the highest weightage among all the regions. We set $h_g = 5$, and $\sigma_g = 1.5$ after cross validation. Finally our total loss for training generator network is given by,
\begin{center}
    $L_G=L_{Gp}+L_{GA}=\frac{1}{N}\sum_x\sum_y p(x,y)\odot [I(x,y)-R(x,y)]^2 -\mathbb{E}_{z \backsim \mathbb{P}_z(z)}[D(G(z))]$
\end{center}

\begin{figure*}[h]
    \centering
    \includegraphics[scale=.5]{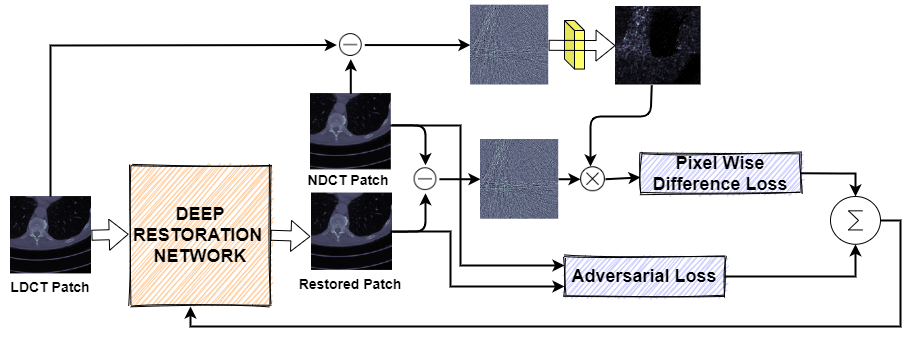}
    \caption{Illustration of our training procedure.}
   \label{fig:my_label}
\end{figure*} 
\section{Experimental Settings}
For validation of our proposed algorithm, we used a publicly available Low dose CT image database by Mayo Clinic\cite{ref_website}. This dataset contains a simulated quarter dose CT scan of 10 patients. Each patient has approximately 250 slices per scan. We used images from 7 patients as the training set and images from 3 patients as the test set. Our test set had more than 550 image pairs. For training, we set the patch size to $120 \times 120$, and randomly cropped ten patches from each slice. We used a total of 21210 patches for training. For comparison, we considered few recent state of the art method, BM3D\cite{ref_article10}, REDCNN\cite{ref_article9}, WAVE-RESNET\cite{ref_article20}, FrameletNet\cite{ref_article7}, WGAN\cite{ref_article15}, CPCE-2D\cite{ref18}. Out of the all above, BM3D don't have any implicit training step. For remaining methods, we trained the network using the same training set and followed the training procedure, as mentioned in the respective papers.
For training our network, we used Adam optimizer with a batch size of 64. The learning rate was initially set to $1e^{-6}$ and $4e^{-6}$ for generator network and discriminator network, respectively, and was set to decrease by a factor of 2 after every 5000 iterations. We trained the network as long there was no improvement in training reconstruction loss for 15000 iterations. We also systematically investigated the role of every module proposed in this study. We considered six variations of our proposed model for comparison as shown in Table \ref{abalation}
\begin{table}[h]
\centering
\resizebox{0.47\textwidth}{!}{%
\begin{tabular}{|c|l|}
\hline
Variant Name & Comment \\ \hline
M1 & Baseline Model without Non Local Module \\ \hline
M2 & Non Local Network trained with Conventional MSE Loss \\ \hline
M3 & Non Local Network trained with Noise Conscious MSE Loss \\ \hline
M4 & Non Local Network trained with Vanilla Discriminator. \\ \hline
M5 & Non Local Network trained with SNMP Discriminator. \\ \hline
M6 & Non Local Network trained with Self Attentive SNMP Discriminator. \\ \hline
\end{tabular}%
}
\caption{SUMMARY OF ALL TRAINED NETWORKS: LOSS FUNCTIONS AND
TRAINABLE NETWORKS.}
\label{abalation}
\end{table}
 \ref{abalation}

\section{Evaluation Matrices}
\begin{figure*}[h]
    \centering
    \includegraphics[scale=.2]{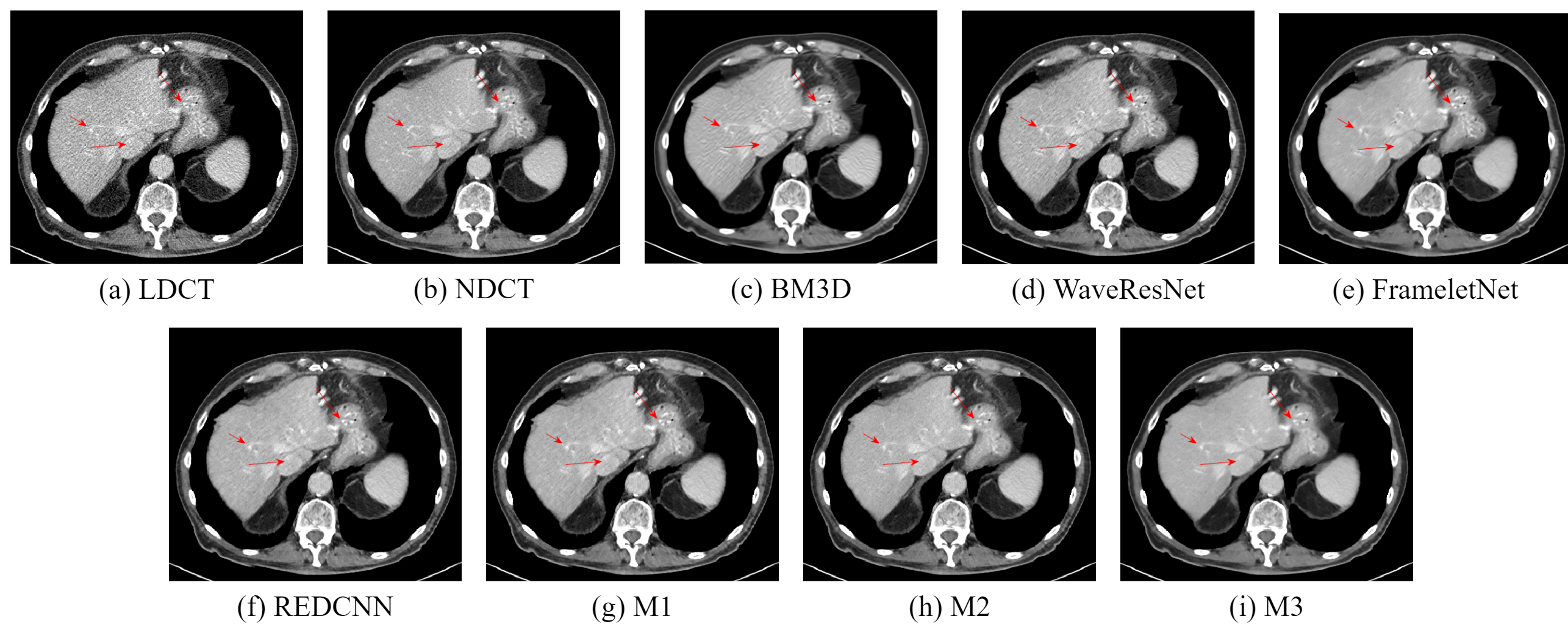}
    \caption{Results of denoising for comparison. (A) LDCT, (B) NDCT, (C) BM3D, (D) WaveRestNet, (E) FrameLetNet, (F) REDCNN, (G) M1, (H) M2, and (I) M3. The arrows indicate three regions to observe the visual effects. The display window is [ $ - 160$, 240] HU for better visualization of the lesion. Viewers are encouraged to zoom in for better view for highlighted area.}
    \label{fig:my_label}
\end{figure*}
In this study we used different matrices for evaluating our different aspect of our proposed module. To make the paper self contained all the evaluation metrics are explained briefly in this section.
\paragraph{\textbf{RMSE}}
Root Mean Square Error (RMSE) measures the pixel distance between the reconstructed images and real images. It is defined as
\[\ RMSE= \sqrt{MSE}\]
where MSE or mean square error is defined as 
\[ MSE=\frac{1}{mn}\sum_0^{m-1}\sum_0^{n-1} \| x(i,j) - g(i,j)\|^2  \]
here $x$ and $g$ are original image and generated image respectively. 
\paragraph{\textbf{PSNR}}
Peak signal to noise ratio (PSNR) is expressed as the ratio between the maximum power present in the signal and the power of the of noise present in the signal. Mathematically it is expressed as 
\[\ PSNR= 20\log_{10}\frac{MAX_I}{\sqrt{MSE}} \]
A high value of PSNR indicates better denoising algorithm.
\paragraph{\textbf{SSIM}}
The Structural Similarity Index (SSIM) gives a measure about the perceptual difference between two similar images. It  infer the given test image is how much similar to the reference image. Higher the value of SSIM indicates more similarity. Mathematically SSIM is defined as follows:
\[\ SSIM=\left(l(x,g)^\alpha . c(x,g)^\beta . s(x,g)^\gamma\right) \]
where, 
$l(x,g)=\frac{2\mu_x\mu_g + c_1}{\mu_x^2+\mu_g^2+ c_1}$, $c(x,g)= \frac{2\sigma_x\sigma_g + c_2}{\sigma_x^2+\sigma_g^2 + c_2}$, $s(x,g)= \frac{\sigma_{xg}+c_3}{\sigma_x\sigma_g + c_3}$, $c_3=c_2/2$
Here, $x$ and $g$ are two image to compare and $\alpha$, $\beta$, $\gamma$ are constant. $l(x,g)$, $c(x,g)$, $s(x,g)$ are luminance, contrast and structure respectively.
\paragraph{\textbf{CNR}}
Contrast to noise ratio measures the visibility of nodules or small details in presence of noise. It is defined as 
\begin{center}
    $CNR=\frac{\mu_1 - \mu_2}{\sqrt{\sigma_1^2 + \sigma_2^2}}$
\end{center}
Where, $\mu_1$, $\mu_2$, $\sigma_1$, and $\sigma_2$ are mean and standard deviation of the region containing nodule and the background respectively. High value of CNR indicates better visibility. 
\paragraph{\textbf{FID}}
Frechet Inception Distance (FID) score is used to evaluate the performance of generative adversarial networks. It uses pre trained Inception v3 model. It is defined as
\begin{center}
 $FID=||\mu_x - \mu_g||_2^2+Tr\left(\sum_x + \sum_g - 2(\sum_x\sum_g)^{0.5}\right)$
\end{center}
Here, $\mu_x$, $\mu_x$, $\sum_x$, $\sum_g$ are the mean and covariance matrix of the last activation layer of inception v3 model for a set of real images($x$) and generated images($g$). A lower value of FID score implies the generated images are more like real images.   
\paragraph{\textbf{TML}} Texture matching loss (TML) is used to measure the  difference between the textural content of reconstructed image and original image. It uses a pre-trained VGG network, and calculates the difference between the correlation of feature of different layers. TML is defined as 
\begin{center}
$TML= \sum_{\forall L}||G\left(\phi_l(g)\right)-G\left(\phi_l(x)\right)||_2^2$   
\end{center}
Here $\phi_l$ is the activation from $l$th layer of the pre-trained VGG network, and Gram Matrix $G(F)=FF^T$.

\section{Result and Discussion}
We organized our results as follows, first we showed the analysis of the proposed non local module to show dependence of denoising performance with neighbourhood size, then we showed the efficacy of our proposed noise aware MSE loss and proposed non local module. And finally the denoising performance of our proposed discriminator and comparison with  other state of art methods.
\begin{figure}
    \centering
    \begin{tikzpicture}
\begin{axis}[
    title={},
    xlabel={Neighbourhood Size $N_i$},
    ylabel={PSNR(dB)},
    xmin=0, xmax=120,
    ymin=32, ymax=33.5,
    xtick={0,10,20,30,40, 60, 80, 100, 120},
    ytick={32.25,32.5,32.75, 33.0,33.25},
    legend pos=north east,
    ymajorgrids=true,
    grid style=dashed,
]

\addplot[
    color=blue,
    mark=square,
    ]
    coordinates {
    (0,32.45)(12,32.85)(20,32.98)(28,32.87)(36, 32.62)(120,32.20)
    };
\end{axis}
\end{tikzpicture}
    \caption{Neighborhood Size ($N_i$) vs. Image denoising performance in terms of PSNR(dB). Neighbourhood size 0 implies network without non local module, and neighbourhood size 120 implies the whole image is used for calculating correlation.}
    \label{fig:neigh}
\end{figure}
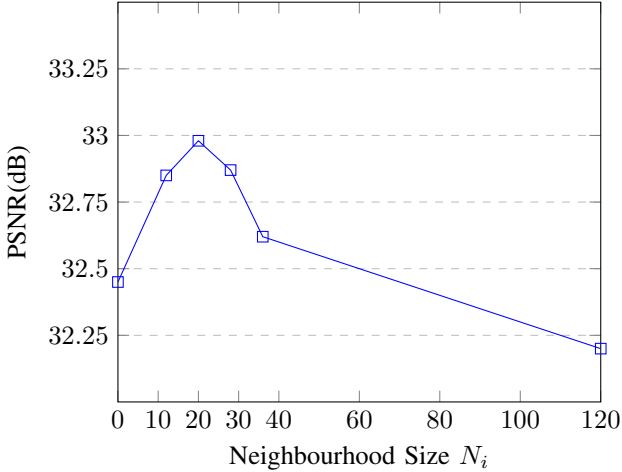

\begin{table*}
\centering
\resizebox{\textwidth}{!}{%
\begin{tabular}{|l|l|l|l|l|l|l|l|}
\hline
Method & BM3D & RED-CNN & WAVE-RESNET & FRAMELT-NET & Proposed M1 & Proposed M2 & Proposed M3 \\ \hline
RMSE   &  11.93    &   10.16      &     10.89        &       10.38      &     10.13        &      9.88       &       \textbf{9.69 }     \\ \hline
PSNR   &   31.70   &     32.45    &     32.07      &      32.28       &     32.46        &     32.76        &       \textbf{32.98 }     \\ \hline
SSIM   &   0.84   &    0.88    &       0.85      &        0.86     &        0.88     &        0.89     &        \textbf{0.905}     \\ \hline
\end{tabular}%
}
\caption{Average PSNR, SSIM, and RMSE of taken over all the slice of the test set. Best results are marked in bold.}
\label{MSE_sota}
\end{table*}

\begin{figure*}[t]
    \centering
    \begin{subfigure}{.24\textwidth}
    \includegraphics[scale = 0.24]{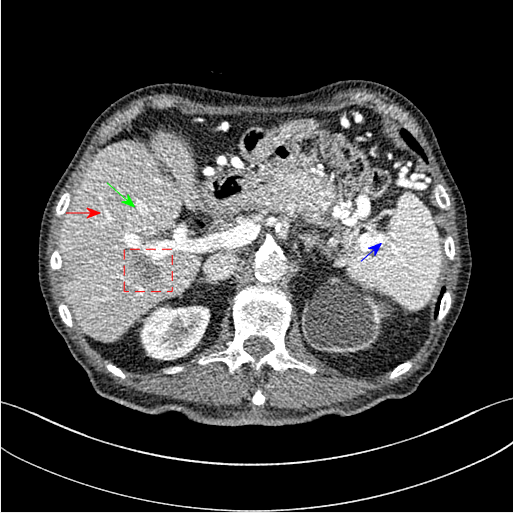}
    \caption{LDCT}
    \end{subfigure}
    \begin{subfigure}{.24\textwidth}
    \includegraphics[scale = 0.24]{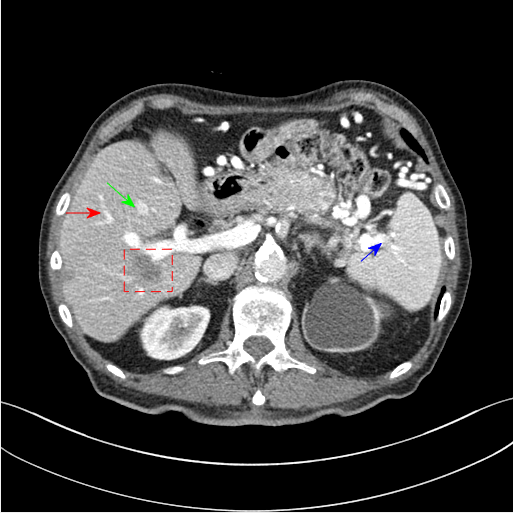}
    \caption{NDCT}
    \end{subfigure}
    \begin{subfigure}{.24\textwidth}
    \includegraphics[scale = 0.24]{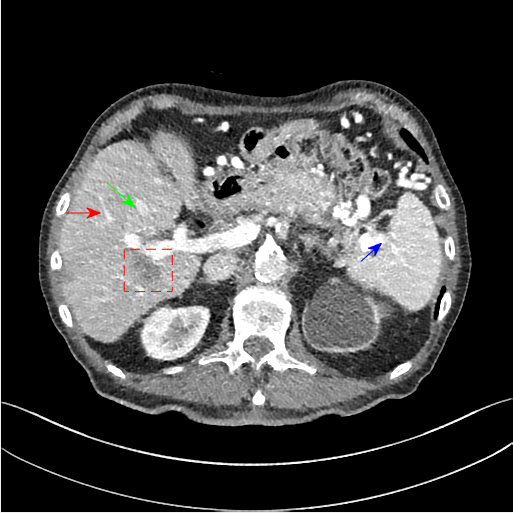}
    \caption{WGAN}
    \end{subfigure}
    \begin{subfigure}{.24\textwidth}
    \includegraphics[scale = 0.24]{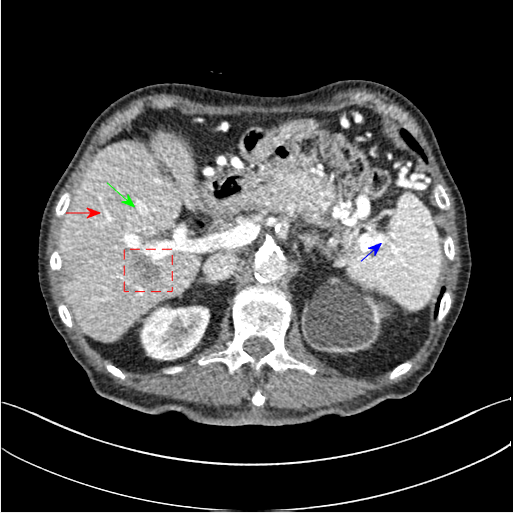}
    \caption{CPCE}
    \end{subfigure}
    \begin{subfigure}{.24\textwidth}
    \includegraphics[scale = 0.24]{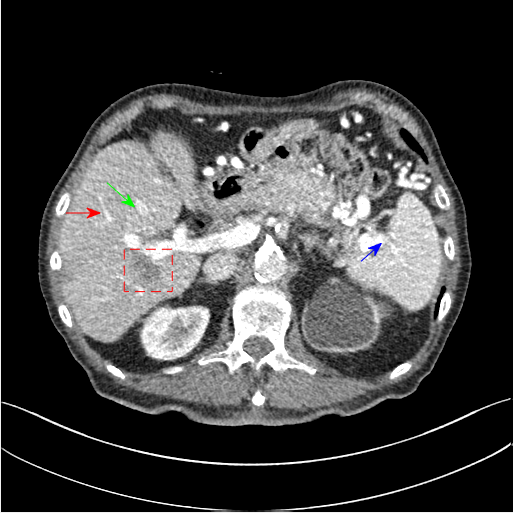}
    \caption{M4}
    \end{subfigure}
    \begin{subfigure}{.24\textwidth}
    \includegraphics[scale = 0.24]{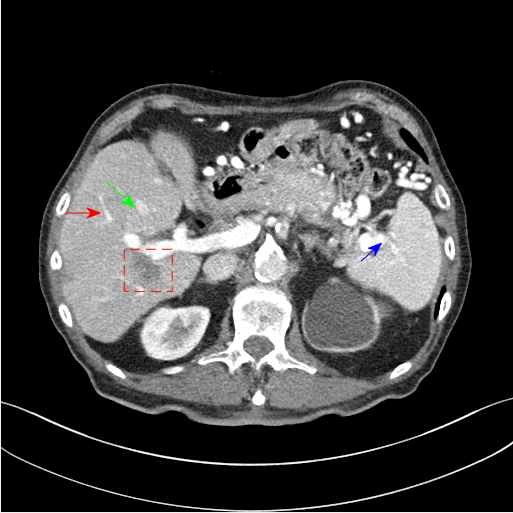}
    \caption{M5}
    \end{subfigure}
    \begin{subfigure}{.24\textwidth}
    \includegraphics[scale = 0.24]{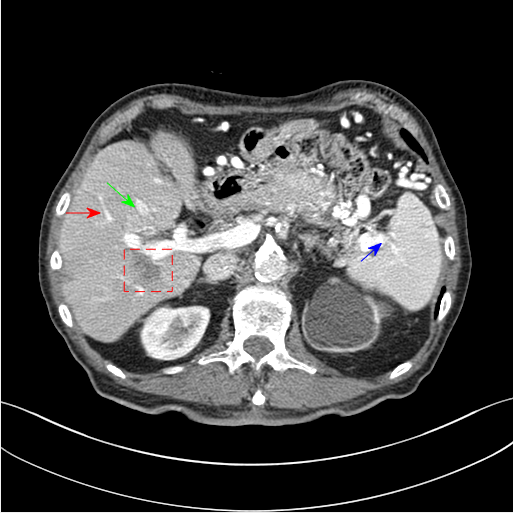}
    \caption{M6}
    \end{subfigure}
    \caption{ Results of denoising for comparison. (A) LDCT, (B) NDCT, (C) WGAN, (D) CPCE, (E) M4, (F) M5, and (G) M6.  The arrows indicate three regions to observe the visual effects. The display window is [ $ - 160$, 240] HU for better visualization of the lesion. Viewers are encouraged to zoom in for better view for highlighted area.}
    \label{fig:my_label}
\end{figure*}

\begin{figure}[t]
    \centering
    \begin{subfigure}{.1\textwidth}
    \includegraphics[scale = 1.1]{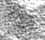}
    \caption{LDCT}
    \end{subfigure}
    \begin{subfigure}{.1\textwidth}
    \includegraphics[scale = 1.1]{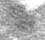}
    \caption{NDCT}
    \end{subfigure}
    \begin{subfigure}{.1\textwidth}
    \includegraphics[scale = 1.1]{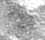}
    \caption{WGAN}
    \end{subfigure}
    \begin{subfigure}{.1\textwidth}
    \includegraphics[scale = 1.1]{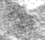}
    \caption{CPCE}
    \end{subfigure}
    \begin{subfigure}{.1\textwidth}
    \includegraphics[scale = 1.1]{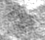}
    \caption{M4}
    \end{subfigure}
    \begin{subfigure}{.1\textwidth}
    \includegraphics[scale = 1.1]{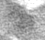}
    \caption{M5}
    \end{subfigure}
    \begin{subfigure}{.1\textwidth}
    \includegraphics[scale = 1.1]{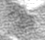}
    \caption{M6}
    \end{subfigure}
    \caption{
Zoomed version of low attenuated lesion outlined in Figure \ref{fig:my_label}. (A) LDCT, (B) NDCT, (C) WGAN, (D) CPCE, (E) M4, (F) M5, and (G) M6. The display window is [ $ - 160$, 240] HU for better visualization of the lesion. Viewers are encouraged to zoom in for better view.}
    \label{fig:crop}
\end{figure}

\begin{figure*}[t]
    \centering
    \begin{subfigure}{.24\textwidth}
    \includegraphics[scale = 0.24]{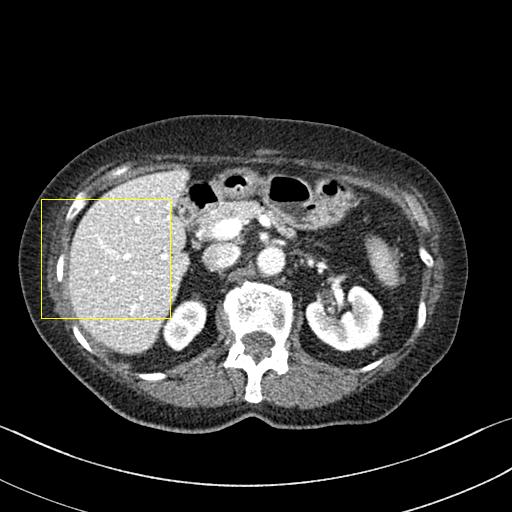}
    \caption{LDCT}
    \end{subfigure}
    \begin{subfigure}{.24\textwidth}
    \includegraphics[scale = 0.24]{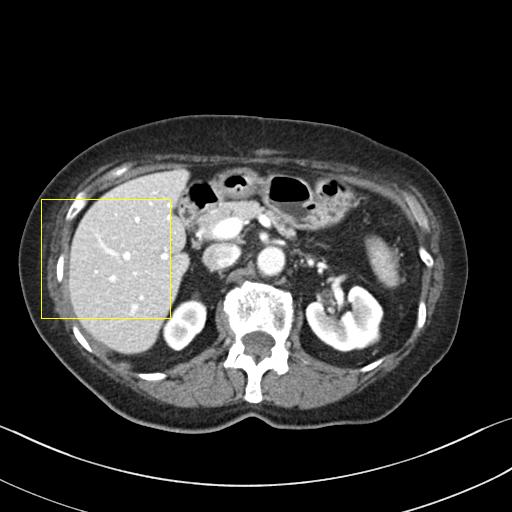}
    \caption{NDCT}
    \end{subfigure}
    \begin{subfigure}{.24\textwidth}
    \includegraphics[scale = 0.24]{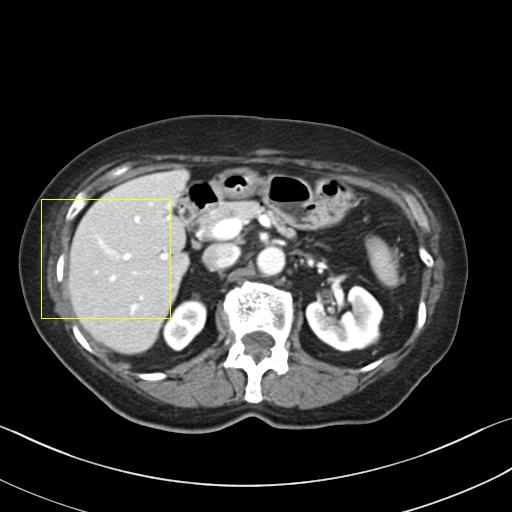}
    \caption{WGAN}
    \end{subfigure}
    \begin{subfigure}{.24\textwidth}
    \includegraphics[scale = 0.24]{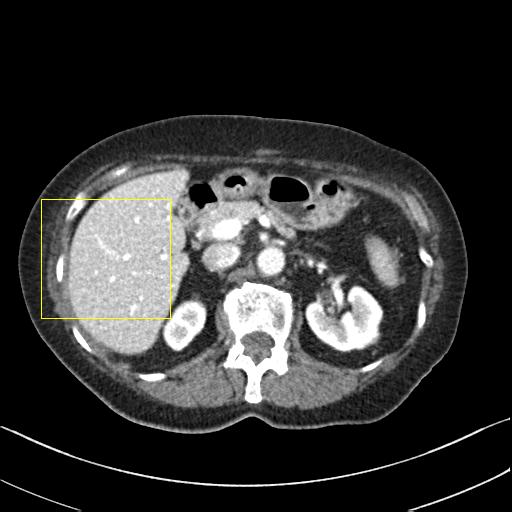}
    \caption{CPCE}
    \end{subfigure}
    \begin{subfigure}{.24\textwidth}
    \includegraphics[scale = 0.24]{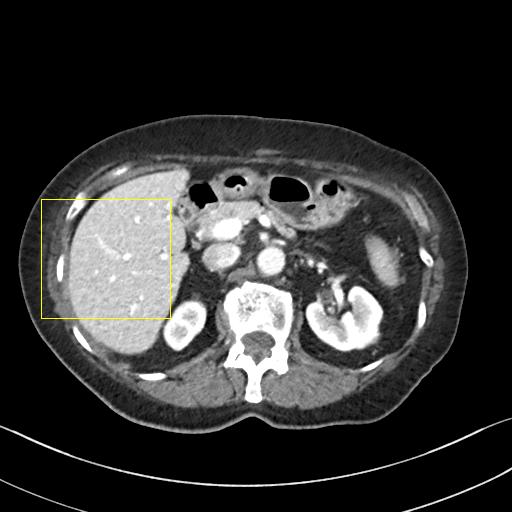}
    \caption{WaveResNet}
    \end{subfigure}
    \begin{subfigure}{.24\textwidth}
    \includegraphics[scale = 0.24]{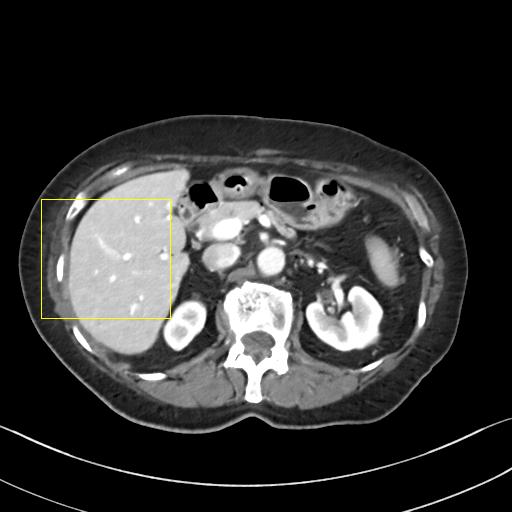}
    \caption{FrameLetNet}
    \end{subfigure}
    \begin{subfigure}{.24\textwidth}
    \includegraphics[scale = 0.24]{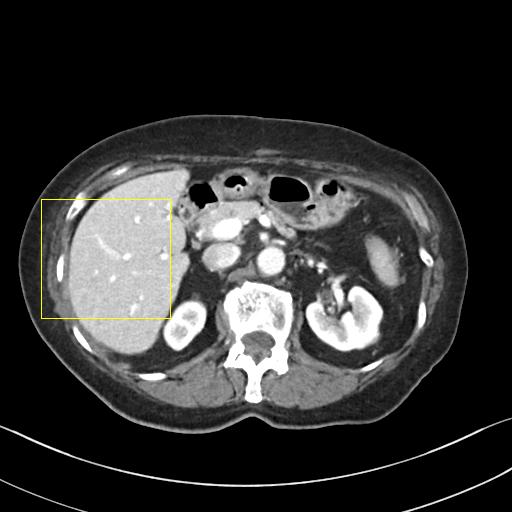}
    \caption{REDCNN}
    \end{subfigure}
    \begin{subfigure}{.24\textwidth}
    \includegraphics[scale = 0.24]{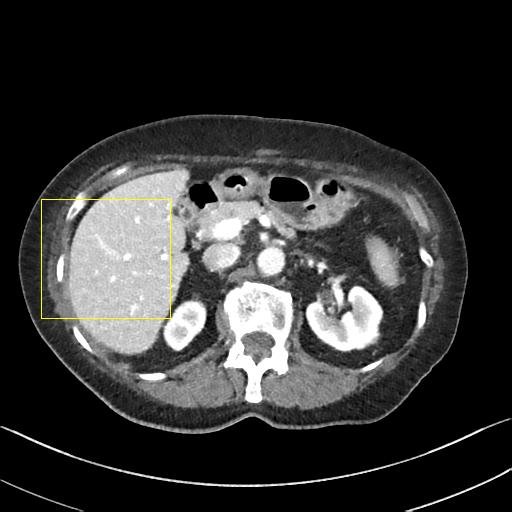}
    \caption{Proposed}
    \end{subfigure}
    \caption{ Result denoising with a low level noisy example. (A) LDCT, (B) NDCT, (C) WGAN, (D) CPCE, (E) WaveResNet, (F) FrameLetNet, and (G) Proposed. The display window is [ $ - 160$, 240] HU for better visualization of the lesion. For low level noise each methods performs adequately. We outlined a region for appreciating the generated texture from different methods. Viewers are encouraged to zoom in for better view for highlighted area.}
    \label{final1}
\end{figure*}

\begin{figure*}[t]
    \centering
    \begin{subfigure}{.24\textwidth}
    \includegraphics[scale = .8]{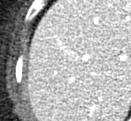}
    \caption{LDCT}
    \end{subfigure}
    \begin{subfigure}{.24\textwidth}
    \includegraphics[scale = .8]{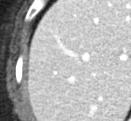}
    \caption{NDCT}
    \end{subfigure}
    \begin{subfigure}{.24\textwidth}
    \includegraphics[scale = .8]{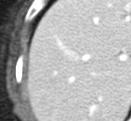}
    \caption{WGAN}
    \end{subfigure}
    \begin{subfigure}{.24\textwidth}
    \includegraphics[scale = .8]{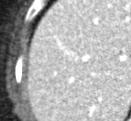}
    \caption{CPCE}
    \end{subfigure}
    \begin{subfigure}{.24\textwidth}
    \includegraphics[scale = .8]{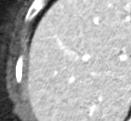}
    \caption{WaveResNet}
    \end{subfigure}
    \begin{subfigure}{.24\textwidth}
    \includegraphics[scale = .8]{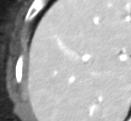}
    \caption{FrameLetNet}
    \end{subfigure}
    \begin{subfigure}{.24\textwidth}
    \includegraphics[scale = .8]{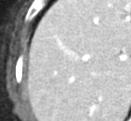}
    \caption{REDCNN}
    \end{subfigure}
    \begin{subfigure}{.24\textwidth}
    \includegraphics[scale = .8]{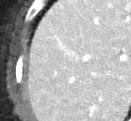}
    \caption{Proposed}
    \end{subfigure}
    \caption{ Zoomed version of the marked region in the Figure \ref{final1}. (A) LDCT, (B) NDCT, (C) WGAN, (D) CPCE, (E) WaveResNet, (F) FrameletNet, (G) REDCNN, and (I) Proposed. The display window is [ $ - 160$, 240] HU for better visualization of the lesion. It can be seen that noise suppression ability of CPCE, WGAN is still not convincing even for low level noise. Although both generated good texture, whereas MSE based methods lost the original texture but suppressed the noise more. However proposed method has good trade off between the texture and noise suppression. Viewers are encouraged to zoom in for better view.}
    \label{final1_crop}
\end{figure*}
\begin{figure*}[t]
    \centering
    \begin{subfigure}{.24\textwidth}
    \includegraphics[scale = 0.24]{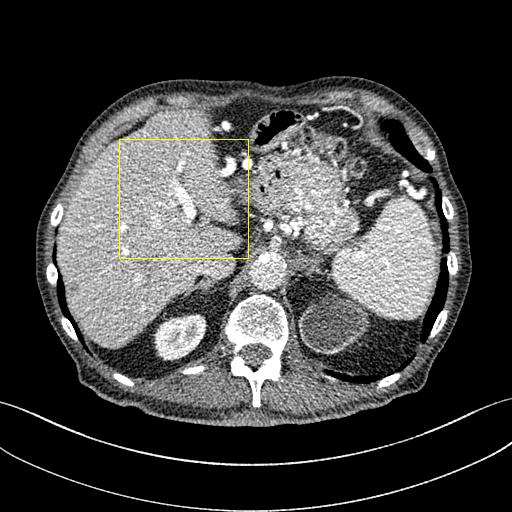}
    \caption{LDCT}
    \end{subfigure}
    \begin{subfigure}{.24\textwidth}
    \includegraphics[scale = 0.24]{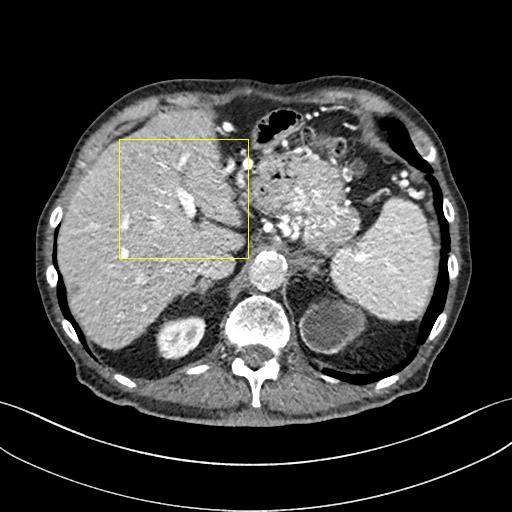}
    \caption{NDCT}
    \end{subfigure}
    \begin{subfigure}{.24\textwidth}
    \includegraphics[scale = 0.24]{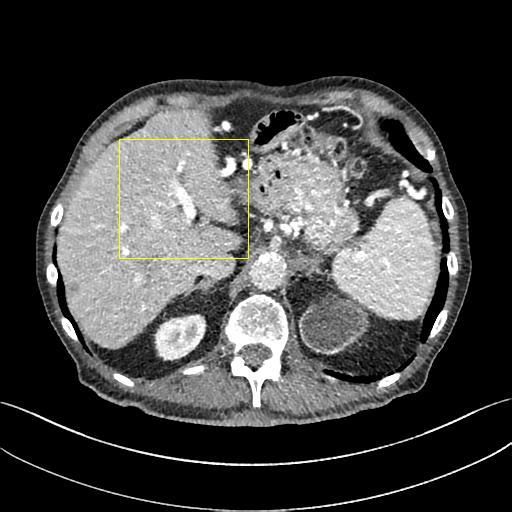}
    \caption{WGAN}
    \end{subfigure}
    \begin{subfigure}{.24\textwidth}
    \includegraphics[scale = 0.24]{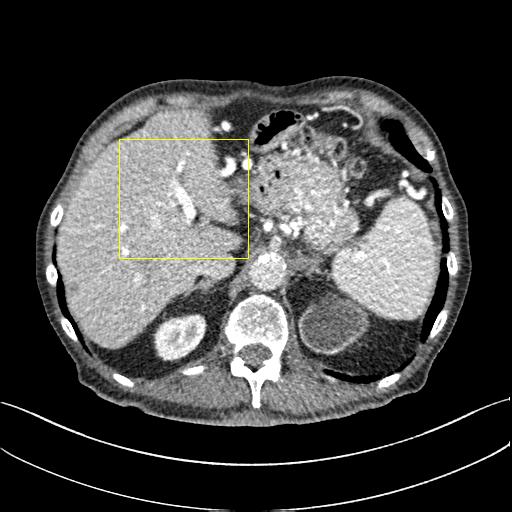}
    \caption{CPCE}
    \end{subfigure}
    \begin{subfigure}{.24\textwidth}
    \includegraphics[scale = 0.24]{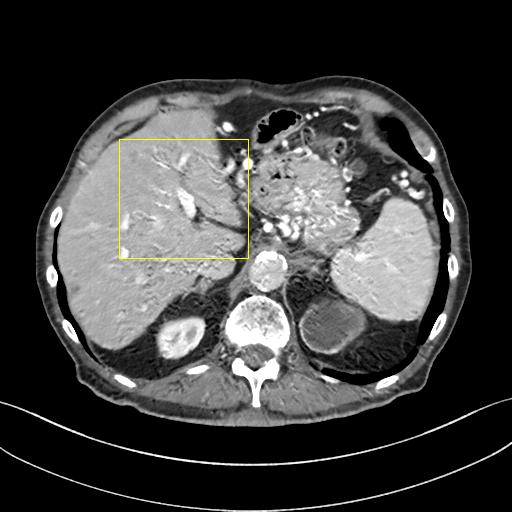}
    \caption{WaveResNet}
    \end{subfigure}
    \begin{subfigure}{.24\textwidth}
    \includegraphics[scale = 0.24]{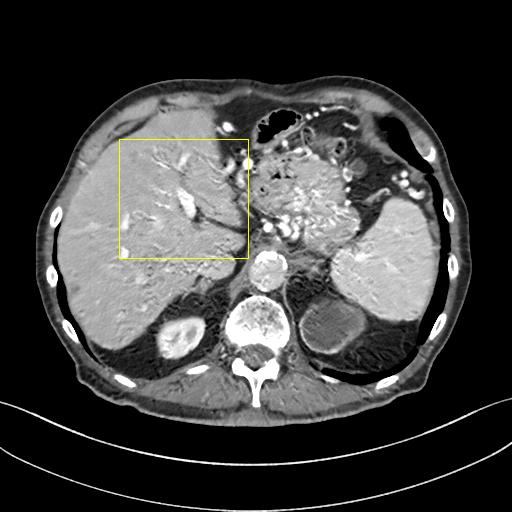}
    \caption{FrameLetNet}
    \end{subfigure}
    \begin{subfigure}{.24\textwidth}
    \includegraphics[scale = 0.24]{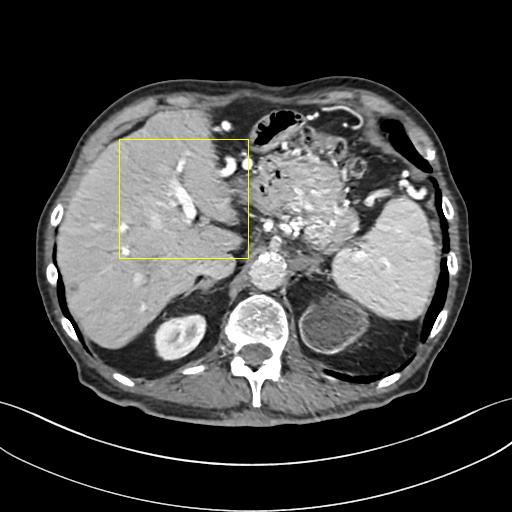}
    \caption{REDCNN}
    \end{subfigure}
    \begin{subfigure}{.24\textwidth}
    \includegraphics[scale = 0.24]{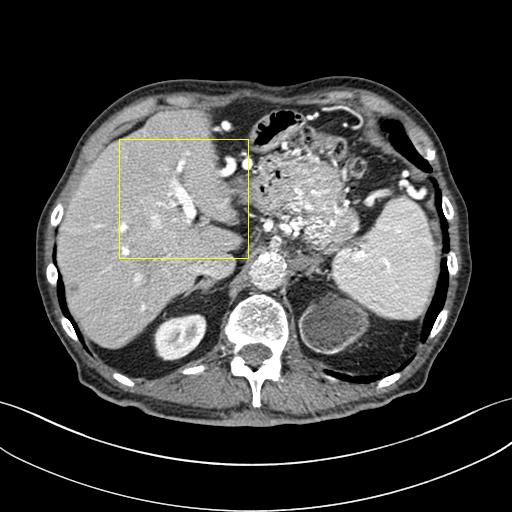}
    \caption{Proposed}
    \end{subfigure}
    \caption{  Result denoising of a high level noisy example. (A) LDCT, (B) NDCT, (C) WGAN, (D) CPCE, (E) WaveResNet, (F) FrameLetNet, and (G) Proposed. The display window is [ $ - 160$, 240] HU for better visualization of the lesion. For low level noise each methods performs adequately. We outlined a region for observing the generated texture from different methods. Viewers are encouraged to zoom in for better view for highlighted area..}
    \label{final2}
\end{figure*}

\begin{figure*}[t]
    \centering
    \begin{subfigure}{.24\textwidth}
    \includegraphics[scale = .8]{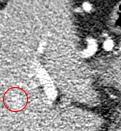}
    \caption{LDCT}
    \end{subfigure}
    \begin{subfigure}{.24\textwidth}
    \includegraphics[scale = .8]{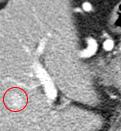}
    \caption{NDCT}
    \end{subfigure}
    \begin{subfigure}{.24\textwidth}
    \includegraphics[scale = .8]{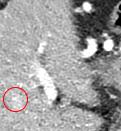}
    \caption{WGAN}
    \end{subfigure}
    \begin{subfigure}{.24\textwidth}
    \includegraphics[scale = .8]{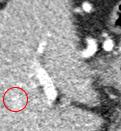}
    \caption{CPCE}
    \end{subfigure}
    \begin{subfigure}{.24\textwidth}
    \includegraphics[scale = .8]{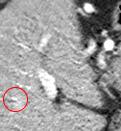}
    \caption{WaveResNet}
    \end{subfigure}
    \begin{subfigure}{.24\textwidth}
    \includegraphics[scale = .8]{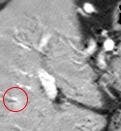}
    \caption{FrameLetNet}
    \end{subfigure}
    \begin{subfigure}{.24\textwidth}
    \includegraphics[scale = .8]{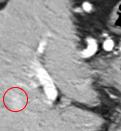}
    \caption{REDCNN}
    \end{subfigure}
    \begin{subfigure}{.24\textwidth}
    \includegraphics[scale = .8]{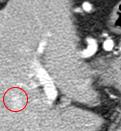}
    \caption{Proposed}
    \end{subfigure}
    \caption{ Zoomed version of the marked region in the Figure \ref{final2}. (A) LDCT, (B) NDCT, (C) WGAN, (D) CPCE, (E) WaveResNet, (F) FrameletNet, (G) REDCNN, and (I) Proposed. The display window is [ $ - 160$, 240] HU for better visualization of the lesion. The outlined a blood vessel which is merely visible in LDCT. Due to insufficient noise suppression in the output of WGAN, and CPCE the blood vessel is non identifiable. In the output of REDCNN and FrameLetNet the vessel is completely missing. The output of proposed method has better visibility of blood vessel. the  Viewers are encouraged to zoom in for better view.}
    \label{final2_crop}
\end{figure*}
Figure \ref{fig:neigh} depicts the dependence of denoising performance with neighbourhood size in terms of PSNR(dB). To represent the performance of the network without using Non Local module we have used neighbourhood size $0$ as a dummy in the diagram. The network size $120$ implies conventional non local module, i.e. the correlation is computed from the whole image similar to self attention mechanism. As seen in the figure, the performance of network is highly sensitive to neighbourhood size. It rapidly increases for small neighbourhoods, and then decreases with increase in neighbourhood size. If the whole image is used then the performance actually degrades due to dissimilarity between the features as discussed in the previous section. The best performance is achieved for neighbourhood size $20$. We have used the neighbourhood size $20$ in our all experiment if not mention anything specifically. The neighbourhood size mentioned here is the neighbourhood size in the full image resolution. As we have down-sampled the image dimension by a factor of 4 before applying non local module, so inside the non local module 5 neighbourhood pixels are used for calculating correlation in case of $N_i=20$.
\par To show the efficacy of our noise conscious MSE loss and proposed non local module, we considered the first three variation of the trained network and compared them with other state of art method, which are trained by MSE minimization. The Table \ref{MSE_sota} depicts the comparative denoising result in terms of PSNR, SSIM, and RMSE. For calculation of metrics we have truncated the values outside the range [-160 240]. As described in Table \ref{abalation} M1 is the baseline network with out non local module and M2 is proposed  non local network. The M2 variant of the trained network has 0.42dB better PSNR than the M1. Inclusion of non local module in the network helps in achieving so. Next the variant M3 yielded 0.2dB better PSNR than M2. The difference between M2 and M3 is only the objective function. M3 is trained with noise conscious MSE loss. So, our proposed loss has assisted the network to focus more on the high signal intensity region, which in return yielded better performance. It can also be seen that our proposed M3 variant has achieved best result among all the state of art methods. Figure 4 gives the visual comparison of different methods. The main objectives of denoising are suppress the noise and recover small details like lesions, nodules, blood vessels and also recover the boundary information and contour of the organs. As pointed by the red arrows the classical method BM3D has recovered the blood vessels efficiently, however noise suppression is not good and also created some artifacts in high noise regions. The FrameLetNet is a iterative reconstruction method, noise suppression of this method is better comparable to other state of the art methods, however the blood vessels have a washed out appearance which has affected the visibility of blood vessels. The output of REDCNN has a better visibility of blood vessels and noise suppression is also adequate. The output of both the M3, and M2 has better visibility than the state of art methods. Moreover output of model M3 has suppress the noise most efficiently among all other methods. Additionally, blood vessels has better visibility and boundary of blood vessels is more defined in M3 than other models as pointed by the red arrows. However all the outputs has a blurry appearance because of MSE loss.
\begin{table*}[]
\centering
\resizebox{\textwidth}{!}{%
\begin{tabular}{|c|l|l|l|l|l|}
\hline
Method Name & PSNR & SSIM & FID & TML & Comment \\ \hline
WGAN & 30.08 & 0.886 & 30.85 & 3.71$e^{-9}$ & Vanilla Discriminator and Perceptual Loss \\ \hline
CPCE & 29.55 & 0.884 & 29.67 & 3.42$e^{-9}$ & Vanilla Discriminator and Perceptual Loss \\ \hline
M4 & 31.81 & 0.905 & 25.69 & 2.12$e^{-9}$ & Vanilla Discriminator and Noise Conscious MSE Loss \\ \hline
M5 & 32.41 & 0.913 & 17.25 & 1.02$e^{-9}$ & SNMP Discriminator and Noise Conscious MSE Loss \\ \hline
M6 & 32.20 & 0.917 & 17.07 & 2.72$e^{-10}$ & Self Attentive SNMP Discriminator and NC MSE Loss \\ \hline
\end{tabular}%
}
\caption{Average PSNR, SSIM, FID, and TML of taken over all the slice of the test set. Best results are marked in bold}
\label{gan}
\end{table*}

\begin{table}[]
\centering
\resizebox{0.2\textwidth}{!}{%
\begin{tabular}{|c|l|}
\hline
Method Name & CNR \\ \hline
NDCT & 0.8273 \\ \hline
LDCT & 0.5200 \\ \hline
WGAN & 0.6275 \\ \hline
CPCE & 0.6011 \\ \hline
M4 & 0.6712 \\ \hline
M5 & 0.7128 \\ \hline
M6 & 0.7962 \\ \hline
\end{tabular}%
}
\caption{CNR of the low attenuated lesion with respect to the background. For calculation we used [$- 160$, 240] window and normalized the values to the range [0,1]}
\label{cnr}
\end{table}
Next, We present the denoising result of self-attentive SNMP discriminator. We consider the last three variations of our proposed model and compared them with WGAN and CPCE-2D methods. Figure \ref{fig:my_label} gives a visual example of the denoising performance of our proposed discriminator and comparison with other states of the art methods. The display window is set to [$-$ 160, 240] for better visualizations of blood vessels and lesions. We identified a few blood vessels and pointed them using arrows of different colors. Also, we outlined a low attenuated lesion using a red color bounding box in the figure. The blood vessel indicated by the green color arrow is almost non-visible in the LDCT image. All the methods have resorted the visibility upto a certain extent; however, the output of M5 and M6 has the most excellent certainty about the vessel. Similar can be seen for other blood vessels also. The zoomed version of the low attenuated the lesion is shown in Figure \ref{fig:crop}. As seen in the figure, the distinctness of the lesions in LDCT is abysmal due to noise. The output of the WGAN and CPCE also suffers the same problem. The output of M4 also doesn't have any defined boundary of the lesion. The contrast between the lesion and the neighborhood is dull.
In M5, the lesion is adequately restored as a result of better noise suppression; however, near the edge, there is some ambiguity about the size and extent of the lesion. In M6, the visibility and also the contrast is similar to the NDCT image. To assess these trades objectively, we used CNR, and the CNR of the lesion with respect to the background is reported in Table \ref{cnr}. CNR can be considered as a representation of visibility in the presence of noise. Our method has outperformed other methods by a considerable margin in terms of CNR, as shown in the table. For calculation, we used [$- 160$, 240] window and normalized the values to the range [0,1]. Next, we report the objective evaluation of our method with other states of the art method in Table \ref{gan}. The main objective of adversarial training is to decrease the distance between the distribution of the resorted images and target images. FID gives a good estimation of the distance between these two distribution. Our proposed model M6 has achieved the lowest FID score, which implies the generated image is more like the real images. The state of the art methods such as WGAN and CPCE has an FID score in the range 30, whereas our proposed method yielded an FID score of 17. The texture of CT gives a lot of information to the radiologist about the disease. We can measure the texture difference by using Texture matching loss or TML. Our proposed method has achieved a TML around ten times lower than the state of the art methods.
The FID score of proposed M5 is similar to the M6; however, by comparing it in terms of TML, M6 has surpassed the M5 by a significant margin. As discussed earlier, the self-attention module enables the discriminator to maintain global consistency. The texture is global information, so self-attentive SNMP has captured it more efficiently than the SNMP discriminator. Lastly, we also compared our model in terms of PSNR and SSIM. Although PSNR and SSIM are not the right metrics for evaluation of the GAN-based model, as they follow the pixel-wise distance. Still, it can be seen; our proposed discriminator yielded good PSNR. Perhaps, it indicates that SN Markovian patch discriminator loss coheres well with per-pixel distance. In the Figure \ref{final1}, and Figure \ref{final2} we have given two more example of denoising. The input of Figure \ref{final1} is less severely affected by the noise. We also marked on region for observing the generated texture of different method. The input of Figure \ref{final2} more strongly  affected by the noise. We also marked on region for observing the small details preservation ability of our method. 
\section{Conclusion}
This paper addressed some common problems associated with deep learning method based LDCT denoising. This work's contribution is as follows: 1) We proposed a distinct nonlocal module, tailored for CT images restoration. To the best of our knowledge, this is the first attempt to utilize neighborhood similarity using deep learning for LDCT denoising. 2) we propose a noise conscious MSE loss for LDCT denoising. The benefit of our proposed loss function is two-fold: first, it gives special attention to the region where noise has distorted the image severely, next it also solves the problem of training deep CNN network with image patches. 3) We also proposed a novel discriminator for LDCT denoising function leveraging self-attention and spectral normalized markovian patch discriminator. 
We evaluated our proposed method on a publicly available dataset of the 2016 NIH-AAPM-Mayo Clinic Low Dose CT Grand Challenge. We have validated each component of our proposed method, both objectively and subjectively. The quantitative comparison of our method with other state of the art methods exhibits our method has crossed all the former methods in terms of PSNR, SSIM, FID, TML by a significant margin. We found the reconstructed image has a praiseworthy CNR, equivalent to the original full dose scan. And subjective comparison showed our method is not only superior in restoring the diagnostic quality but also the produced texture similar to full-dose scans. For future work, we will consider incorporating the neighborhood similarity over the adjacent slices adopting 3D nonlocal operations.  

\end{document}